%% file: main.tex
\def\year{2021}\relax
\documentclass[letterpaper]{article} 
\usepackage{aaai21}  
\usepackage{times}  
\usepackage{helvet} 
\usepackage{courier}  
\usepackage[hyphens]{url}  
\usepackage{graphicx} 
\def\UrlFont{\rm}  
\usepackage{graphicx}  
\usepackage{natbib}  
\usepackage{caption} 
\usepackage{booktabs}
\usepackage[inline]{enumitem}
\usepackage[colorinlistoftodos,textsize=tiny]{todonotes}
\usepackage[ruled,vlined]{algorithm2e}
\usepackage[switch]{lineno}
\newcommand{\filip}[1]{\textcolor{blue}{[\textbf{FI:} #1]}}
\newcommand{\jf}[1]{\textcolor{red}{[\textbf{JF:} #1]}}

\newcommand{\km}[1]{\textcolor{violet}{[\textbf{KM:} #1]}}

\newcommand{\atomic}{\texttt{ATOMIC}}
\newcommand{\conceptnet}{\texttt{ConceptNet}}
\newcommand{\wordnet}{\texttt{WordNet}}
\newcommand{\wikidata}{\texttt{Wikidata}}
\newcommand{\CSKG}{\texttt{Commonsense Knowledge Graph}}
\newcommand{\cskg}{\texttt{CSKG}}
\newcommand{\visgen}{\texttt{VisualGenome}}

\newcommand{\csqa}{\texttt{CommonSenseQA}}
\newcommand{\siqa}{\texttt{SocialIQA}}
\newcommand{\piqa}{\texttt{PhysicalIQA}}
\newcommand{\anli}{\texttt{aNLI}}
\newcommand{\winogrande}{\texttt{WinoGrande}}

\newcommand{\cut}[1]{}

\newcommand{\paperidvar}{1059}

\begin{document}
%
\title{Knowledge-driven Data Construction for Zero-shot Evaluation\\ in Commonsense Question Answering}

\author{
Kaixin Ma\textsuperscript{\rm 1},
Filip Ilievski\textsuperscript{\rm 2},
Jonathan Francis\textsuperscript{\rm 1,3},\\
Yonatan Bisk\textsuperscript{\rm 1},
Eric Nyberg\textsuperscript{\rm 1},
Alessandro Oltramari\textsuperscript{\rm 3}\\
}
\affiliations{
\textsuperscript{\rm 1}Language Technologies Institute, School of Computer Science, Carnegie Mellon University\\
\textsuperscript{\rm 2}Information Sciences Institute, Viterbi School of Engineering, University of Southern California\\
\textsuperscript{\rm 3}Human-Machine Collaboration, Bosch Research Pittsburgh\\
\{kaixinm, jmf1, ybisk, ehn\}@cs.cmu.edu, ilievski@isi.edu, alessando.oltramari@us.bosch.com
\vspace{2mm}
}

\maketitle

\begin{abstract}
\begin{quote}
Recent developments in pre-trained neural language modeling have led to leaps in accuracy on commonsense question-answering benchmarks. However, \cut{whereas large-capacity neural systems are able to model individual datasets,} there is increasing concern that models overfit to specific tasks, without learning to utilize external knowledge or perform general semantic reasoning. In contrast, zero-shot evaluations have shown promise as a more robust measure of a model's general reasoning abilities\cut{, as models that achieve state-of-the-art performance on individual datasets suffer significant performance-degradation under zero-shot evaluation on new, but similar tasks}. In this paper, we propose a novel neuro-symbolic framework for zero-shot question answering across commonsense tasks. Guided by a set of hypotheses, the framework studies how to \cut{unify and} transform various pre-existing knowledge resources into a form that is most effective for pre-training models. We vary the set of language models, training regimes, knowledge sources, and data generation strategies, and measure their impact across tasks. Extending on prior work, we devise and compare four constrained distractor-sampling strategies. We provide empirical results across five commonsense question-answering tasks with data generated from five external knowledge resources. We show that, while an individual knowledge graph is better suited for specific tasks, a global knowledge graph brings consistent gains across different tasks. In addition, both preserving the structure of the task as well as generating fair and informative questions help language models learn more effectively.

\end{quote}
\end{abstract}

\input{1-intro}

\input{2-related-work.tex}

\input{3-approach}

\input{4-results}

\input{5-discussion}

\input{6-conclusion}

\clearpage
\section{Acknowledgement}
We would like to thank all of the anonymous reviewers for their valuable feedback. We also thank Ehsan Qasemi and Soumyaroop Nandi for participating in our human annotation study. This material is based upon work sponsored by the DARPA MCS program under Contract No. N660011924033 with the United States Office Of Naval Research.

\bibliography{aaai.bib}
\clearpage
\appendix

\section{Hyperparameters}
\label{appendix:params}
\subsection{Model training} In all experiments that involve training, we used learning rate $1e^{-5}$, batch size $32$, max sequence length $128$, weight decay $0.01$, adam epsilon $1e^{-6}$, $\beta_1 = 0.9$, $\beta_2 = 0.98$ and warm-up proportion $0.05$, margin $1.0$. For MLM training with RoBERTa, We only mask non-stop words in head or tail and we set the masking probability as 0.5 for \texttt{ATOMIC} and 0.3 for \texttt{CWWV}.
\subsection{Tuned parameters}
We also tested smaller margins for training (0.2, 0.5) and we observe that margin 1.0 works slightly better. 
\subsubsection{AFLite}
In our experiments, we set N=64, $\tau$=0.75, $k_1$=1/50 of $|Trn|$, $k_2$=1/50 of $|Dev|$ and O = 1/5 of $|Trn|$.
\section{AFLite algorithm}
\label{appendix:AFLite}
\begin{algorithm}
\SetAlgoLined
 \KwIn{$Trn$, $Dev$, ensemble size N, cutoff sizes $k_{1}$,$k_{2}$, threshold $\tau$, target size O }
 
 \While{$|Trn| > O$ }{
  \For{s in Trn+Dev}{Initialize Prediction P(s) = $\emptyset$}
  \For{i=1, ... N}
    {Random Partition $Trn$ into $U$, $V$ s.t $|U|$ = O \\
    Train a linear classifier F on $U$ \\
    \For{s in V+Dev} {Add F(s) in P(s)} 
    }
  \For{s in Trn+Dev}{Acc(s) =  $\frac{|p \in P(s) s.t. p = y|}{|P(s)|}$}
  Select top-$k_{1}$ samples $S_1$ in $Trn$  s.t. Acc(s) $> \tau$\\
  Select top-$k_{2}$ samples $S_2$ in $Dev$ s.t. Acc(s) $> \tau$\\
  $Trn$ = $Trn$ - $S_1$\\
  $Dev$ = $Dev$ - $S_2$\\
  \lIf{$|S_1| < k_{1}$ }{break}
  }
 \KwOut{filtered sets $Trn'$ and $Dev'$}
 \caption{AFLite}
\end{algorithm}

\section{Computing resources}
We run our experiments on servers with Intel(R) Core(TM) i7-7820X CPU @ 3.60GHz
(1 CPU, 8 physical cores per CPU, total 16 logical CPU units) and with 125GB RAM. For GPUs, We used Nvidia RTX 2080Ti and Nvidia Titan RTX. For libraries, we used Pytorch 1.2.0, transformers 3.0.2 and sentence-transformer 0.3.4. 
\end{document}

%% file: 1-intro.tex
\section{Introduction}
\label{introduction}

Common sense is key to efficient communication in everyday situations, as it enables natural language understanding through contextual reasoning. 
Machine question answering (QA) benchmarks, like \siqa~\cite{sap-etal-2019-social} and \piqa~\cite{Bisk2020}, are effective {\it behavioral} tests of commonsense reasoning in machines, each focusing on different capabilities. Answering a question in \siqa~might require the knowledge that readers typically prefer heroes over villains in fantasy novels; whereas, in \piqa, the knowledge that metal stools can break windows, because windows are made of glass and metal is a more enduring material than glass. 
Although such tasks had been traditionally difficult for machines, recent developments in pre-trained neural language modeling have led to leaps in accuracy|closing the gap between human and machine performance to single-digit percentage points.\footnote{For example (accessed 4 August, 2020): \url{https://leaderboard.allenai.org/socialiqa/submissions/public}} However, due to increasing concern that large-capacity neural systems are modeling individual datasets, rather than learning how to perform logical reasoning or to utilize external knowledge effectively~\cite{mitra2019exploring}, focus is shifting to alternative training and evaluation strategies.
In particular, \textit{zero-shot evaluation} shows promise as an efficient measure of model generalisability across tasks~\cite{Shwartz2020UnsupervisedCQ,\cut{Banerjee2020SelfsupervisedKT,}li-etal-2020-harvesting}.
Here, models are trained and validated on task $\mathbf{A}$, and tested on a different task $\mathbf{B}$, without access to $\mathbf{B}$'s training data or labels. This leads state-of-the-art models from individual tasks to falter, sometimes  by as much as a 50\% decrease in performance~\cite{Shwartz2020UnsupervisedCQ}. \cut{While \textit{few-shot} evaluation (allowing a few training examples from $\mathbf{B}$) can help, the gaps to the fully-supervised training and human-level performance remain large. }

Repositories of commonsense knowledge, like \conceptnet~\cite{10.5555/3298023.3298212} and \atomic~\cite{DBLP:conf/aaai/SapBABLRRSC19}, can be beneficial for commonsense QA, especially when little or no training data is available.
%
\cut{\km{this gap refers to the gap between zero-shot and human, but the following papers are not addressing this issue in particular}}
Enriching the training data with \conceptnet~and \atomic~has been shown~\cite{ma-etal-2019-towards,mitra2019exploring} to improve accuracy on datasets \textit{derived} from these graphs: \csqa~\cite{talmor-etal-2019-commonsenseqa} and \siqa. 
Knowledge bases (KBs) can be used to generate question-answer pairs and distractors automatically, in order to test a model's reasoning ability~\cite{petroni2019language,richardson2019does} or provide additional supervision~\cite{ye2019align,Yang2020GDAUGGD}. 
\cut{\cut{possible solution to these issues\cut{\km{maybe we want to clarify what these issues refer?}}; this is typically seen as a}way of making a given contextual representation more complete \cite{rao-daume-iii-2019-answer}, providing additional supervision \cite{klein2019learning, dhingra-etal-2018-simple}, or of generating tests for models' reading comprehension~\cite{du-etal-2017-learning}. Refinement techniques such as adversarial filtering~\cite{bras2020adversarial} may further improve their quality, by using embedding similarity in order to automatically distill informative questions for a language model. \cut{\km{this sentence seems isolated here}}}
While KBs have been shown to help in a zero-shot transfer setting recently~\cite{Banerjee2020SelfsupervisedKT}, no comprehensive study exists on the relation between various knowledge, its usage method, and neural models for zero-shot transfer across commonsense tasks. Moreover, while adversarial filtering techniques~\cite{bras2020adversarial} improve the quality of a manually created question set, their impact on automatically generated questions from a variety of KBs has not been investigated yet.

In this paper, (1) we compile a set of hypotheses and design a novel neuro-symbolic framework that investigates the dependency between knowledge sources, question generation techniques, language model (LM) variants, and tasks. Our framework leverages a wide range of KBs, covering visual, social, and concept-based knowledge, to pre-train LMs for zero-shot evaluation on multiple-choice commonsense QA tasks.
(2) Recognizing that the aspect of question generation is especially understudied, we expand on prior work to devise and test four distractor-sampling strategies for effective question generation.\cut{ for non-extractive commonsense QA tasks.}  
We analyze their impact on model performance across tasks, conditioned on model class and (pre-)training regime, and show that generating questions that are simultaneously fair and informative is difficult but beneficial for LM pre-training.
(3) We determine which combination of knowledge graphs (KGs), data construction/training, and architectures is most effective\cut{for transfer to downstream tasks} and can utilize appropriately rich contexts across five tasks.\cut{We provide a systematic evaluation of the effects of on downstream zero-shot model performance for  benchmarks.} We observe that diversifying knowledge generally improves performance, under the condition of it being aligned with the task, and that preserving the structure of the task is desired.
(4) We make our code and resulting datasets available to the community to facilitate future research in this direction.\footnote{\url{https://github.com/Mayer123/HyKAS-CSKG}} \

%% file: 2-related-work.tex
\section{Related Work}


\subsection{Knowledge Injection}


\cut{Commonsense capabilities on a variety of topics (e.g., social or physical understanding) are commonly evaluated by multiple-choice QA benchmarks, like 
\siqa~
and \piqa~
}
Strong performance on standard multiple-choice QA benchmarks, like \siqa~and \piqa, has been achieved by fine-tuning a task-specific prediction layer, placed atop pre-trained LMs, such as BERT \citep{devlin-etal-2019-bert}, RoBERTa \citep{Liu2019RoBERTaAR}, and GPT \citep{radford2019language}. 
As shown by \citet{ma-etal-2019-towards} and \citet{mitra2019exploring},
combining neural methods with structured background knowledge from \conceptnet, \wordnet~\citep{10.1145/219717.219748}, and \atomic~ works well for commonsense datasets that have been partially derived from these resources, such as \siqa~ and \csqa. Here, the structured knowledge, formalized as lexicalized task-targeted evidence paths, is injected into an LM, either
\cut{The injection of such structured evidence into a LM can be done with }via an attention mechanism \citep{bauer-etal-2018-commonsense} or through an auxiliary training objective \citep{10.1145/3357384.3358165}. Graph and relation networks can also be used to score answer candidates, by informing the graph structure with data from LMs~\citep{lin-etal-2019-kagnet,10.1007/978-3-030-32233-5_2}. 
Finally, complete KGs can be incorporated directly in training by introducing additional modeling objectives, to teach a model about general commonsense regardless of the task at hand~\citep{peters-etal-2019-knowledge,levine-etal-2020-sensebert,Liu2020KBERTEL,zhang-etal-2019-ernie,talmor2020teaching}. This line of work resembles our approach of including background knowledge in a general, task-agnostic way; however, it still relies on the task training data and has generally not been tested in a zero-shot regime.


\subsection{Generating Commonsense Questions and Answers}

\citet{richardson2019does} use links in \wordnet~to generate question-answer pairs, 
then leverage the resulting dataset to evaluate language models. ~\citet{petroni2019language} prompt the skills of language models by sentences instead of questions, generated from sources like \texttt{ConceptNet} and SQuAD~\cite{rajpurkar2016squad}.
Previous works have generated synthetic QA sets to complement existing training data. \cut{model training, albeit mostly in addition to existing training data.} \citet{ye2019align} proposed an `align-mask-select' method to generate questions using \texttt{ConceptNet} and Wikipedia. \citet{kocijan-etal-2019-surprisingly} constructed a large set of pronoun resolution questions using Wikipedia sentences. \citet{Yang2020GDAUGGD} generate QA pair and distractors using generative models. 
Regarding zero-shot evaluation, the Self-Talk model of~\cite{Shwartz2020UnsupervisedCQ} generates clarification prompts based on a template prefix, which are leveraged to elicit knowledge from another LM, which is used jointly with the original context and question to score each answer candidate. 
Given a task context, one can use COMET~\citep{Bosselut2019DynamicKG}, a generative model trained on commonsense KGs, to generate background knowledge statements, and to compute scores for each answer candidate based on the context, question, and generated knowledge.
\citet{Banerjee2020SelfsupervisedKT} pre-train the LM with three representation learning functions which aim to complete a knowledge triple given two of its elements. These functions jointly compute the distance for each answer candidate. The ambition of this paper is to provide a comprehensive framework for such prior efforts on zero-shot QA with KGs. By covering a wider set of KGs, question generation techniques, and tasks, we can systematically investigate the effect of using different KGs, generation methods, and techniques across tasks.

%% file: 3-approach.tex
\section{Zero-Shot QA Framework}

Given a natural language question $Q$, and $n$ possible answers $A_1, ..., A_n$, the task \cut{of a system (human or machine)} is to select the most probable single answer $A$. We refer to the remaining $n-1$ possible answers: $D_1, ..., D_{n-1}$ as distractors. In a zero-shot QA evaluation mode, the system has no access to the task training or development data. 
We assume a setup where the system is pre-trained once and then applied across different tasks in a zero-shot manner. Our zero-shot evaluation framework addresses this task by variants of pre-training an LM on an artificial QA set, created from KG data. Next, we describe its covered tasks, sources of knowledge, question generation strategies, LM techniques, and training regimes, in turn.\cut{ of our framework.} \cut{We describe each of these dimensions, in turn.}





\subsection{Synthetic QA Generation}

\begin{figure}[!t]
\includegraphics[scale=0.23]{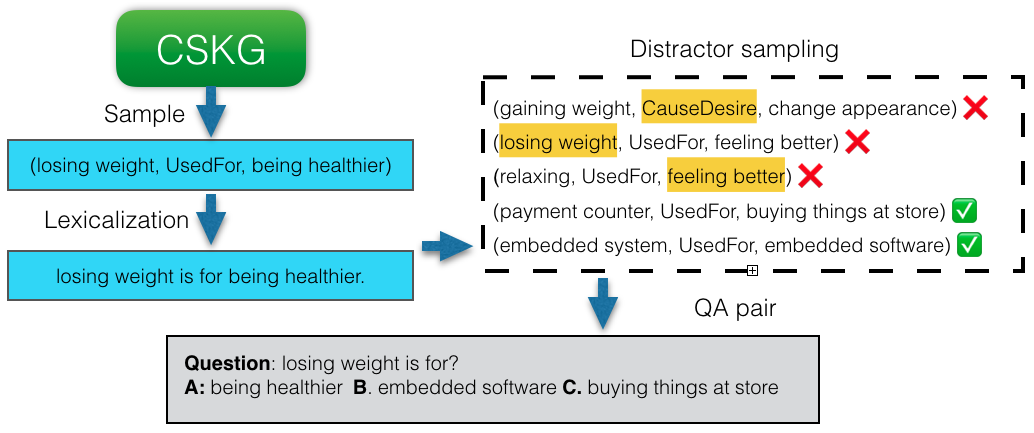}
\caption{An illustration of our question generation pipeline.\cut{\filip{we need to explain the example here...}}}
\label{fig:generation}
\end{figure}

We generate questions, answers, and distractor options from five KGs: \atomic, \conceptnet, \wordnet, \visgen~\cite{krishna2017visual}, and \wikidata~\cite{vrandevcic2014wikidata}, found in the \CSKG~(\cskg)~\cite{ilievski2020consolidating}. Notably, \texttt{ATOMIC} differs from the other KGs in two ways: 1) its relations have a different focus than those of the other sources; and 2) its node labels are longer and formalized as templates. Due to these considerations, we prepare two sets of QA sets: one based on \texttt{ATOMIC} and one based on the remaining four knowledge sources. Figure \ref{fig:generation} illustrates our question generation pipeline. 

\subsubsection{Data partitions} 
\texttt{ATOMIC} expresses pre- and post-states for events and their participants with nine relations. Its head nodes are events, whereas the tail nodes are either events or attributes. Its nodes have two particularities: 1) irrelevant parts of the node text are replaced with blanks (`\_'); and 2) references to fictional agents are indicated with special tokens (e.g., \textit{PersonX}). We follow the \siqa's \texttt{ATOMIC} train/dev/test splits, to ensure that the facts of the dev and test partitions are excluded in training. 

Our second partition, \texttt{CWWV}, covers three other KGs in \cskg~\cut{\km{is this the first mention of CSKG?}}that express commonsense facts between concepts: \conceptnet, \wordnet, and \wikidata. We use them jointly to generate questions, and we enrich them with additional distractors from \visgen. Treating these four sources as a single one is enabled by their \cskg~mapping to a single set of relations, defined by \conceptnet. We focus on 14 semantic relations that are grounded on strong psycholinguistic and pragmatic evidence \cite{murphy2003semantic}, like \texttt{/r/Causes} and \texttt{/r/HasPrerequisite}. Since there is no pre-defined train/dev/test split for \cskg, we randomly sample 5\% of generated questions as development set, while the other 95\% are used for training, to maximize the coverage of the knowledge.  

\subsubsection{Generating questions and answers}
\cut{\todo{Can we have a nice figure diagram/cartoon that illustrates this process (including the negative samples)?  I'm concerned this bit is crucial but confusing}}
If a triple ($h$, $r$, $t$) has an associated sentence, we directly employ it for question generation; otherwise, we generate a sentence in a lexicalization step, using a set of pre-defined templates. Next, we generate the question $Q$ by removing the tail of the sentence, and extract this tail as the correct answer, $A$. 
Here, we ensure that there is no token overlap between the head and the correct answer. For \texttt{ATOMIC}, we: 1) compare the keyword tokens instead of all tokens, in order to avoid stopwords; and 2) the agent templates (e.g., `PersonX') are replaced with randomly sampled gender-neutral names from a pre-defined set.
For \texttt{CWWV}, we filter out questions where either the head or the tail are not common concepts or they are named entities. We use corpus frequency as a proxy for commonness,\footnote{\url{https://pypi.org/project/wordfreq/} (Accessed 9 Sept. 2020)} while named entities are filtered by removing all concepts whose labels start with a capital letter. 

\begin{table}[t]
  \begin{center}
    \small{
    \caption{Generated questions from \atomic~ (top) and \texttt{CWWV} (bottom). (*) denotes the correct answer.}
    \label{tab:example}
    \begin{tabular}{@{}l@{}} 
     \toprule
      Question: Robin takes the fifth. As a result, Robin wanted to\\
      A1: go to the cinema. \\
      A2: withhold information. (*)\\
      A3: hear what they think. \\
    \midrule
      Question: losing weight is for \\
      A1: being healthier. (*) \\
      A2: embedded software. \\
      A3: buying things in store. \\
    \bottomrule
    \end{tabular}
    }
  \end{center}
\end{table}
\subsubsection{Generating negative samples (distractors)}

We seek to generate distractor options that satisfy two criteria: \textit{informativeness} and \textit{fairness}. Namely, a good distractor has semantic relatedness with the context (informative), while being relatively easy to discriminate from the correct answer (fair).
We create the pool of distractors $D$ for every sample as follows:
\begin{enumerate*}
    \item The distractor candidates are the tails of knowledge triples ($h'$, $r'$, $t'$) with the same relation $r' = r$, randomly sampled from the KGs. This would ensure that the distractors can fill the same semantic role as the correct answer. 
    \item The head $h'$ of the sampled triples does not have non-stop word overlap with $h$. 
    \item The distractor tail $t'$ is not part of the correct answer set, i.e., there exist no triples, ($h$, $r$, $t'$)
\end{enumerate*}. 
\cut{These rules would help ensure that the distractors are semantically similar to the answer, while not being correct answers, thus making them simultaneously challenging and fair.}
Considering the example in Figure \ref{fig:generation}, the triple \textit{(gaining weight, CausesDesire, change appearance)} will be filtered out by rule (1), \textit{(losing weight, UsedFor, feeling better)} will be ruled out by both (2) and (3), and \textit{(relaxing, UsedFor, feeling better)} will be ruled out by (3). Here, we replace any references to fictional \texttt{ATOMIC} agents in the distractors with the same names used in the question. We then randomly select two distractors ($D_1$, $D_2$) from $D$. We refer to this distractor pooling strategy as \textit{random}, and propose three alternative strategies in the next Section. 

Example questions with each partition are shown in Table~\ref{tab:example}. For \atomic, this procedure generates 535K QA pairs for training and 60K for development. For \texttt{CWWV}, the training set contains 157K and the dev set has 8K QA pairs.

\subsection{Distractor Sampling}
Existing data generation procedures are likely to introduce annotation artifacts in datasets \cite{zellers-etal-2019-hellaswag,sakaguchi2019winogrande}. Models may exploit these artifacts to achieve spuriously strong performance during training, at the expense of degradation in robustness.
To generate more challenging QA pairs from KGs and to alleviate potential biases in our synthetic sets, we test two other distractor sampling strategies in addition to the \textit{random} strategy: 1) we select distractors that are as similar as possible to the answer, while being under a certain threshold (\textit{adv-answer}); and 2) we select distractors that are as similar as possible to the question, while being under a certain threshold (\textit{adv-question}). Here we define similarity of two nodes to be their proximity in the embedding space, measured by cosine similarity. The intuition is that, by generating more challenging QA pairs for the models, we could achieve better generalization across tasks. We use the RoBERTa sentence embedding model~\cite{reimers-2020-multilingual-sentence-bert} to compute embeddings for all KG nodes. For these two strategies, we set an upper bound on the similarity score to avoid unfair distractors, i.e., paraphrases of the correct answer. Based on manual observations, we set their distractor similarity upper bound to be 0.6 for \texttt{CWWV} and 0.4 for \texttt{ATOMIC}.

\subsubsection{Sample filtering}
Besides these distractor sampling strategies, we test another condition (3), where we select the distractors randomly, but only keep the questions whose distractors are sufficiently challenging at training time (\textit{adv-filter}). The intuition is that QA pairs generated using the aforementioned methods might still be too easy for the models, thus we would like to only keep the most challenging subset to train our models. We employ the AFLite algorithm \cite{sakaguchi2019winogrande} for our purpose. Given a \textit{train} and \textit{dev} split of our synthetic QA set, we use 5\% of the \textit{train} set to finetune a RoBERTa model with a classification head (4\% training, 1\% validation). These 5\% are discarded from \textit{train} after this step. We then compute the fixed embeddings for the remaining 95\% of \textit{train} and the entire \textit{dev}, denoted as \textit{Trn} and \textit{Dev}. Next, we feed \textit{Trn} and \textit{Dev} along with their labels to the AFLite algorithm, which iteratively filters out easy examples using an ensemble of linear classifiers. Finally, we retain (101K training, 11K dev) samples for \texttt{ATOMIC} and (29K training, 1.5K dev) samples for \texttt{CWWV} subset. The details of AFLite can be found in the appendix.

\cut{we partition the remaining 95\% of \textit{train} into \textit{train-sub} and \textit{train-dev} set, and train a linear classifier taking the computed embeddings as the input on the \textit{train-sub}. (d) We make prediction on \textit{train-dev} and \textit{dev} sets. (e) We repeat step (c) and (d) for K times and train a new classifier on a different partition each time. (f) We filter out examples that has prediction accuracy greater than $\tau$. (g) While the remaining size of \textit{train} is larger than \textit{Target} or the newly discarded sample is greater than \textit{k}, repeat step (e) and (f).}\cut{\filip{the algorithmic steps are a bit hard to follow. Perhaps we should turn this into an Algorithm with pseudo code.}  \filip{i propose we move parameter settings to implementation}}

\subsection{Language Models}
We consider 2 types of language models: auto-regressive language models and masked language models (MLM). Specifically, we use GPT-2 and RoBERTa to select the best answer candidate. Given a context $C$, a question $Q$, and a list of answer options ($A_1, A_2 ... $), we concatenate $C$ and $Q$ with each answer option to build input sequences ($T_1, T_2 ... $). We also use templates to convert a sequence $T$ into a natural language sentence following \cite{Shwartz2020UnsupervisedCQ}. 
For example, we transform the sequence: \textit{[$C$] What will X want to do next? [$A_i$]}  into: \textit{[$C$], as a result, X want to [$A_i$]}. The score $S$ for the resulting sequence using an auto-regressive LM is computed as follows:
\begin{equation}
    \mathrm{S_{LM}}\left(T \right)=-\frac{1}{n} \sum_{i=1}^{n} \log P\left(t_{i} \mid t_{1} \ldots t_{i-1}\right)
\end{equation}
where $n$ is the number of tokens in the sequence and $P$ is the conditional probability provided by the LM. To evaluate MLMs, we mask out one token at a time and compute its loss~\cite{Zhou2020EvaluatingCI}. We repeat this process for every token in the sequence. The final MLM score is:
\begin{equation}
    \mathrm{S_{MLM}}\left(T \right)=-\frac{1}{n} \sum_{i=1}^{n} \log P\left(t_{i} \mid \ldots t_{i-1},  t_{t+1} \ldots \right)
\end{equation}
The predicted option is the one with the lowest score.

\subsubsection{LM Finetuning}
In the typical model architecture for finetuning LM for multiple-choice tasks, a linear layer is added on top of the LM encoder to predict the answer. The model inputs are separated by a model-specific delimiter. However, as this architecture introduces randomly initialized parameters, it may not be able to fully utilize the pre-trained weights~\cite{tamborrino-etal-2020-pre}. 
Instead, we re-use the GPT-2 and RoBERTa with LM head for finetuning.
By keeping the model intact, we can reuse the same converting templates and scoring functions.
To train the model, given the scores computed for each answer candidate $S_1, S_2, ... S_m$, we use the marginal ranking (MR) loss defined as:
\begin{equation}
    \mathcal{L}=\frac{1}{m} \sum_{i=1 \atop i \neq y}^{m} \max \left(0, \eta-S_{y}+S_{i}\right)
\end{equation}
Here, $\eta$ represents the margin and $y$ is the index of the correct answer. For a MLM model, the computation cost for the scoring function scales quadratically with the input length. To make the training more efficient, we only mask out non-stop tokens in the head and tail nodes. 



\subsubsection{Training Regimes}
In order to disentangle the contribution of the KGs from the structure of the QA pairs, we consider different training methods for augmentation of language models with KGs. Specifically, we compare marginal ranking (MR) training with masked language modeling (MLM) training. For MLM, we directly concatenate the question and the correct answer in our synthetic QA set and then train RoBERTa on the these sentences using the MLM objective.\cut{\filip{not sure why is this a separate section}}

\begin{table*}[t]
  \begin{center}
    \caption{
    Zero-shot evaluation results with different combinations of models and knowledge sources, across five commonsense tasks. \texttt{CSKG} represent the combination of \texttt{ATOMIC} and \texttt{CWWV}. We run our experiments three times with different seeds and report average accuracy with 95\% confidence interval. SMLM (*) used OMCS for CSQA, ROCStories \cite{mostafazadeh-etal-2016-corpus} for aNLI and ATOMIC for SIQA as knowledge resources. 
    \cut{Model results under zero-shot inference conditions. \jf{Many different dimensions are captured in this table, and we should carefully describe the distinctions: pre-training vs. not pre-training, ATOMIC vs. CSKG, unsupervised vs. supervised, MLM vs. `new'. We need to emphasise that all the methods that are currently in the `unsupervised (pre-trained)' section (that specifically use \atomic~or \texttt{CSKG}) are ``ours". We should consider adding results for pre-training on a competing QA-generation approach.}. \jf{Explain ``Majority", here in the caption.}. The asterisk (*) denotes the test set accuracy.}
    }
    \label{tab:zero-shot}
    \begin{tabular}{@{}l@{\hspace{5pt}}c@{\hspace{7pt}}c@{\hspace{7pt}}c@{\hspace{7pt}}c@{\hspace{7pt}}c@{\hspace{7pt}}c@{}} 
     \toprule
    \bf Model & \bf KG &  \bf aNLI  & \bf CSQA & \bf PIQA & \bf SIQA & \bf WG \\
      \midrule
       Majority & - & 50.8 & 20.9 & 50.5 & 33.6 & 50.4 \\
      GPT2-L & - & 56.5  & 41.4 & 68.9 & 44.6 & 53.2 \\
      RoBERTa-L & - & 65.5  & 45.0 &  67.6 & 47.3 & 57.5 \\
      Self-talk \hfill \cite{Shwartz2020UnsupervisedCQ} & - & -  & 32.4 & 70.2 & 46.2 & 54.7\\
      COMET-DynaGen \hfill \cite{Bosselut2019DynamicKG} & \texttt{ATOMIC} & - & - & -  & 50.1 & - \\
      SMLM     \hfill \cite{Banerjee2020SelfsupervisedKT} & \texttt{*} & 65.3 & 38.8 & - & 48.5 & - \\
      \midrule
      GPT2-L (MR) & \texttt{ATOMIC} & $59.2(\pm 0.3)$ & $48.0(\pm 0.9)$ & $67.5(\pm 0.7)$ & $53.5(\pm 0.4)$ & $54.7(\pm 0.6)$\\ 
      GPT2-L (MR) & \texttt{CWWV} & $58.3(\pm 0.4)$ & $46.2(\pm 1.0)$ & $68.6(\pm 0.7)$ & $48.0(\pm 0.7)$ & $52.8(\pm 0.9)$\\
      GPT2-L (MR) & \texttt{CSKG} & $59.0(\pm 0.5)$ & $48.6(\pm 1.0)$ & $68.6(\pm 0.9)$ & $53.3(\pm 0.5)$ & $54.1(\pm 0.5)$\\
      RoBERTa-L (MR) & \texttt{ATOMIC} & $\bf 70.8(\pm 1.2)$  &  $64.2(\pm 0.7)$ &  $72.1(\pm 0.5)$ & $63.1(\pm 1.5)$ & $59.6(\pm 0.3)$\\
      RoBERTa-L (MR) & \texttt{CWWV} & $70.0(\pm 0.3)$ &  $\bf 67.9(\pm 0.8)$ & $72.0(\pm 0.7)$  & $54.8(\pm 1.2)$ & $59.4(\pm 0.5)$ \\
      RoBERTa-L (MR) & \texttt{CSKG} & $70.5(\pm 0.2)$ & $67.4(\pm 0.8)$ & $\bf 72.4(\pm 0.4)$ & $\bf 63.2(\pm 0.7)$ & $\bf 60.9(\pm 0.8)$ \\
      \midrule
      \it RoBERTa-L (supervised) & - & 85.6  & 78.5 & 79.2 & 76.6 & 79.3 \\
      \midrule
      \it Human & - & 91.4 & 88.9 & 94.9 & 86.9 & 94.1 \\
    \bottomrule
    \end{tabular}
  \end{center}
\end{table*}

\subsection{Tasks}
\cut{\filip{I wouldn’t put parts of the framework after ‘tasks’}}
We select commonsense tasks based on two criteria. Firstly, we strive to cover a diverse set of tasks, both in terms of their format (question answering, pronoun resolution, natural language inference), as well as their type of knowledge (e.g., social or physical knowledge). Secondly, we prefer larger task datasets that are manually constructed. For this reason, we do not include datasets like COPA~\cite{gordon-etal-2012-semeval}, or HellaSwag~\cite{zellers-etal-2019-hellaswag}. We opt for the following five task datasets:

\subsubsection{1. Abductive NLI (aNLI)}~\cite{bhagavatula2019abductive} is posed as a natural language inference task. Given the beginning and the ending of a story, the task is to choose the more plausible hypothesis out of two options. The dataset consists of nearly 170k entries.

\subsubsection{2. CommonsenseQA (CSQA)}~\cite{talmor-etal-2019-commonsenseqa} evaluates a broad range of common sense aspects. 
Each entry contains a question and 5 answer candidates. The questions are crowdsourced based on a subgraph from \texttt{ConceptNet}. The answer candidates combine \texttt{ConceptNet} nodes with additional crowdsourced distractors. 

\subsubsection{3. PhysicalIQA (PIQA)}~\cite{Bisk2020} is a two-choice question answering dataset which focuses on physical reasoning. Given a question, the system (or human) is asked to pick the more plausible out of two possible continuations.

\subsubsection{4. SocialIQA (SIQA)}~\cite{sap-etal-2019-social} is a question-answering dataset which requires reasoning about social interactions. Each entry contains a context, a question, and 3 answer candidates. The context is derived from the \texttt{ATOMIC} knowledge graph, the questions are generated based on nine templates (corresponding to the relations in \texttt{ATOMIC}), and the answers are crowdsourced.



\subsubsection{5. WinoGrande (WG)}~\cite{sakaguchi2019winogrande} contains 44 thousand pronoun resolution problems. Each entry consists of a context description with an emphasized pronoun, and two options are offered as its possible references.

\section{Experimental Setup}
\subsection{Baselines}
We compare our results with the following baselines. \textit{Majority} answers each question with the most frequent option in the entire dataset. \textit{`Vanilla' versions of the language models} are used in order to understand the impact of further tuning. Here we directly uses the LMs to score the QA pairs without any finetuning. We also show the results of other unsupervised systems that leverage KGs: \textit{Self-talk}, \textit{COMET-}\textit{DynaGen}, and \textit{SMLM}. To indicate the upper bound of this work, we include results of a supervised fine-tuned RoBERTa system and of human evaluation.

\subsection{Implementation}
For the LM baselines, we directly load the weights from the Transformers library~\cite{Wolf2019HuggingFacesTS} and evaluate on the downstream tasks. The finetuned LMs are trained for a single epoch on our synthetic QA set. For Adv-filter, \cut{In the AFLite condition, }we train the models for 5 epochs to compensate for less training data. We use our synthetic dev set to select the best model.\cut{ checkpoints.} We describe other hyper-parameters used and computing infrastructure in the appendix.

\subsection{Hypotheses}
\label{ssec:hypotheses}

Based on individual prior findings and understanding of different components of our framework, we put forward a set of hypotheses which will be validated in our experiments:

\begin{enumerate}[label=H\arabic*]
    \item \textit{RoBERTa would have better performance than GPT-2.} This is in line with prior findings that RoBERTa has the advantage of bi-directional context~\cite{Zhou2020EvaluatingCI}.
    \item \textit{Pre-training a language model with artificially created question-answer sets enhances zero-shot performance.} This is also supported in previous study about unsupervised QA \cite{li-etal-2020-harvesting}
    \item \textit{The impact of more knowledge depends on the alignment between KGs and the task}, partial evidence for which is provided by~\cite{ma-etal-2019-towards,mitra2019exploring}.
    \item \textit{Adding diverse knowledge (from different KGs) improves performance}. This is the initial motivation behind the creation of \texttt{CSKG}~\cite{ilievski2020consolidating}, but has not been investigated in detail.
    \item \textit{When selecting negative samples for a question, it helps to use an adversarial strategy that ensures the question is not trivial for a language model.} H5 is inspired by adversarial filtering, which has not been investigated in detail for automatically-generated questions and across KGs.
    \item \textit{Preserving the task structure when generating synthetic data leads to better accuracy.}
    This is implicitly assumed in prior data augmentation work \cite{kocijan-etal-2019-surprisingly}.
    \item \textit{The automatically created questions are notably easier for humans than they are for machines} - a general assumption made by commonsense task creators and typically correct for any existing, human-generated benchmark.
\end{enumerate}

\begin{table*}[]
    \centering
    \caption{Comparison of different QA generation strategies.}
    \begin{tabular}{lcccccc}
         \toprule
         \bf RoBERTa-L & \bf Strategy & \bf aNLI  & \bf CSQA & \bf PIQA & \bf SIQA & \bf WG \\
      \midrule
       \hspace{1em} +\texttt{ATOMIC} & Random & $\bf 70.8(\pm 1.2)$  &  $\bf 64.2(\pm 0.7)$ &  $72.1(\pm 0.5)$ & $\bf 63.1(\pm 1.5)$ & $59.6(\pm 0.3)$\\
       \hspace{1em}  +\texttt{ATOMIC} & Adv-answer & $70.4(\pm 0.8)$ & $62.3 (\pm 0.9)$ & $\bf 72.6(\pm 1.8)$ & $61.6(\pm 0.3)$ & $60.5(\pm 0.5)$  \\
       \hspace{1em} +\texttt{ATOMIC} & Adv-question & $70.8(\pm 0.6)$ & $55.6(\pm 0.9)$ & $70.6(\pm 0.8)$ & $51.6(\pm 0.8)$ & $58.5(\pm 0.3)$ \\
        \hspace{1em} +\texttt{ATOMIC} & Adv-filter & $68.6(\pm 1.8)$ & $46.4(\pm 1.5)$ & $67.9(\pm 1.1)$ & $51.8(\pm 1.2)$ & $\bf 60.8(\pm 0.6)$ \\
        \midrule
       \hspace{1em}  +\texttt{CWWV} & Random & $\bf 70.0(\pm 0.3)$ &  $ 67.9(\pm 0.8)$ & $72.0(\pm 0.7)$  & $\bf 54.8(\pm 1.2)$ & $59.4(\pm 0.5)$ \\
       \hspace{1em} +\texttt{CWWV} & Adv-answer & $69.5(\pm 1.1)$ & $\bf 68.5(\pm 0.8)$ & $\bf 72.7(\pm 0.3)$ & $53.8(\pm 0.6)$ & $\bf 60.7(\pm 0.7)$ \\
       \hspace{1em} +\texttt{CWWV} & Adv-question & $68.3(\pm 2.3)$ & $60.9(\pm 2.3)$ & $69.6(\pm 0.6)$ & $47.0(\pm 2.0)$ & $59.0(\pm 1.4)$ \\
       \hspace{1em} +\texttt{CWWV} & Adv-filter & $69.7(\pm 0.7)$ & $64.7(\pm 2.3)$ & $72.0(\pm 1.3)$ & $50.1(\pm 1.0)$ & $59.4(\pm 1.4)$ \\
      \bottomrule
    \end{tabular}
    \label{tab:generation}
\end{table*}

%% file: 4-results.tex
\section{Results}
We evaluate various combinations of: knowledge sources, question generation strategies, LMs, training regimes, and tasks. We use accuracy as a metric. All our experiments are performed in a zero-shot setting, i.e., the models do not leverage the official training data of the task. We report results on the dev sets of these tasks, as the official test sets are not publicly available. We note that, since we did not use the tasks' dev sets for hyperparameter tuning or checkpoint selection, the dev sets can be used effectively as test sets. 

\subsection{Main Results}
Table 2 shows that GPT-2 and RoBERTa outperform the majority baseline by a large margin on all tasks, indicating that the LMs have already learned relevant knowledge during pretraining. Despite being a smaller model, RoBERTa outperforms GPT-2 on 4 out of 5 tasks without pretraining, and on all tasks when pretraining over different synthetic QA sets. This shows the advantage of leveraging bi-directional context, and confirms our hypothesis H1. As expected (H2), training RoBERTa on our \texttt{ATOMIC} or \texttt{CWWV} synthetic sets brings notable performance gain on all 5 tasks. We observe that models trained on \texttt{ATOMIC} sets have a large advantage on SIQA compare to models trained on \texttt{CWWV}, while \texttt{CWWV} brings advantage on the CSQA task. This is not surprising as these two tasks are derived from \texttt{ConceptNet} and \texttt{ATOMIC}, respectively. The difference between \texttt{ATOMIC} and \texttt{CWWV} on the remaining three tasks is relatively small. This supports our hypothesis H3: knowledge alignment is crucial for obtaining better performance.

Training on the combined question set (\texttt{CSKG}) is mostly able to retain the best of its both partitions. Training on \texttt{CSKG} leads to best performance on three out of five tasks, showing that a global commonsense resource is able to bring consistent gain across different tasks. This supports our hypothesis H4: adding more diverse knowledge is beneficial for language models. Finally, even with this knowledge, we recognize that there is still a large gap between our model's accuracy and that of the supervised RoBERTa model.

\cut{\subsection{Few-shot Evaluation}
Following previous work, we also experimented our model in few-shot condition. Here we randomly sampled 2400 examples in the training set (~8\% of training set) and continue trained our model. The results are shown in table \ref{tab:few-shot}. Our model pretrained on the synthetic data is able to outperform the baseline models and previous approach, showing that our proposed approach could be useful for the low-resource setting as well. Moreover, it is worth noting that our model in few-shot setting is able to achieve similar performance to that of the RoBERTa model using the entire training set. }

\subsection{Comparison of QA Generation Strategies}


Table~\ref{tab:generation} shows the results with different sampling strategies, thus addressing H5. The best performing adversarial algorithm, \textit{Adv-answer}, yields comparable accuracy to the \textit{random} strategy, revealing that distractors sampled with a more sophisticated strategy are not necessarily more informative for the LMs. \textit{Adv-question} and \textit{Adv-filter} typically lead to declines in accuracy. Considering \textit{Adv-question}, this could be due to the similarity of the distractors to the question, which might guide the model to learn to pick the most dissimilar candidate as the correct answer, which is an artifact of our question generation and cannot be expected to work well for downstream tasks. \cut{Finally, the results for \textit{Adv-filter} dropped significantly on some tasks. }
Our manual inspection of the remaining questions prefered by \textit{Adv-filter} indicates that many questions are unfair, as some distractors are also correct answers, which is a consequence of the incompleteness of the KGs. \cut{, even with our strict rule-based filter, the distractor pool may still contain some unfair distractors.} 
Adv-filter prioritizes these questions as they are ``difficult'' for LMs, however, training on them might teach the LM incorrect knowledge and harm downstream accuracy. 

\subsection{Comparison of Training Regimes}
Table~\ref{tab:regimes} presents results with two different training regimes. In comparison to the baseline without additional training, MLM training on \texttt{ATOMIC} only improves on the SIQA task, and harms on the rest. With \texttt{CWWV}, it brings large gain on CSQA and small improvements on SIQA and WG. At the same time, marginal ranking training on either question set consistently outperforms MLM training by a large margin, suggesting that preserving the task structure is beneficial in addition to the question content and validating H6.

\subsection{Difficulty of the Synthetic QA Sets}
Ideally, the generated question-answer pairs should be challenging for the models but easy for humans to solve (H7).
Here, we probe this hypothesis by assessing the difficulty of our synthetic QA sets both by humans and `vanilla' LMs. 
We evaluated both models on the dev sets of our synthetic data. 
For human evaluation, we randomly sample 50 questions from \texttt{ATOMIC} and 50 questions from \texttt{CWWV}. A total of five researchers were asked to first \textit{provide the correct answer}, then \textit{rate the question difficulty}. For the latter, the annotator chose between easy, moderate, hard, or non-sensical - as a guideline, nonsensical questions have unfair distractors and cannot be easily understood. Following this procedure, we obtained three judgments for each question. 

The inter-annotator agreement on selecting the correct answer is 0.62 using Fleiss Kappa score, which is substantial agreement. The Kripendorf alpha \cite{krippendorff04} for rating question difficulty is 0.35, which is fair agreement. The results of the baseline LMs and human performance (Table~\ref{tab:difficulty}) show that the \texttt{ATOMIC} subset presents a harder challenge for both models, as well as for humans. Overall, the results support our hypothesis H7: the synthetic questions are relatively easy for humans to solve and much harder for models. However, the annotation pointed to several directions for improving the synthetic QA sets. A number of questions generated from \texttt{ATOMIC} are ungrammatical, which makes them harder to understand, while some questions from \texttt{CWWV} were rated as unfair. For example, all answer options for the question \textit{A person can} are valid: (a) \textit{cost several thousand dollars} (b) \textit{expressing boredom} (c) \textit{check snow level}. As discussed earlier, this is due to the incompleteness of our KGs, and the current lack of understanding on how to generate fair, yet informative, distractors. 

\begin{table}
  \begin{center}
    \caption{Comparison between MLM and MR training.}
    \label{tab:regimes}
    \begin{tabular}{@{}ll@{\hspace{7pt}}c@{\hspace{7pt}}c@{\hspace{7pt}}c@{\hspace{7pt}}c@{\hspace{7pt}}c@{}} 
     \toprule
      RoBERTa-L & Train & aNLI & CSQA & PIQA & SIQA & WG\\
        \midrule 
       baseline & - & 65.5  & 45.0 &  67.6 & 47.3 & 57.5 \\
      + \texttt{ATOMIC} & MLM & 62.9 & 43.8 & 65.8 & 53.9 & 55.5 \\
      + \texttt{ATOMIC} & MR & 70.8 & 64.2 & 72.1 & 63.1 & 59.6 \\
      + \texttt{CWWV} & MLM & 65.3 & 57.3 & 67.2 & 49.3 & 59.4 \\
      + \texttt{CWWV} & MR & 70.0 & 67.9 & 72.0 & 54.8 & 59.4 \\
    \bottomrule
    \end{tabular}
  \end{center}
\end{table}

%% file: 5-discussion.tex
\section{Discussion}

\subsection{Towards a Commonsense Service}

The overarching pursuit of this paper is to understand whether generating artificial QA sets with KGs improves the zero-shot QA performance of LMs. We observe a consistent leap in accuracy across tasks, LMs, knowledge sources, and question generation strategies. While these accuracies are notably below supervised LM accuracies, they might further improve by architectural improvements of the LMs, knowledge sources with wider coverage and stronger semantics, and well-tuned scoring functions.\footnote{For example, scoring sequences of tokens by a language model might improve the performance of LMs \cite{tamborrino-etal-2020-pre}.} 
In addition, despite its complexity and diversity, commonsense knowledge (unlike knowledge on entities and events, which changes rapidly) is largely static and evolves slowly over time, thus making the dataset-specific finetuning unnecessary in theory.
A natural question arises: can we build a sufficiently reliable, general commonsense service, by pretraining a LM on a rich set of questions covering a wide spectrum of knowledge types?

\subsection{Impact of Knowledge}

In general, we observed that using knowledge from a wider set of sources is beneficial. However, on \anli~and \csqa, the best accuracy was obtained with a subset of all questions. This could be due to the kinds of knowledge covered: 
\begin{enumerate*}
\item \anli~focuses on expectations of agents and events, for which \texttt{ATOMIC} is directly useful, whereas the other KGs might be delusive;
\item \csqa~mostly requires knowledge about properties of objects (e.g., function or appearance) which is the focus of \texttt{ConceptNet},~\texttt{Wikidata}, and \texttt{VisualGenome}, but not of \texttt{ATOMIC}.
\end{enumerate*}
This indicates a tension between H3 and H4: while more knowledge often helps, it might not be the case when the task and the knowledge are not well-aligned. Our current understanding of the dimensions of commonsense knowledge in knowledge sources and benchmarks is limited, and would benefit from further study.

\begin{table}
  \begin{center}
    \caption{LM and human accuracy on our synthetic QA sets.}
    \label{tab:difficulty}
    \begin{tabular}{lcc} 
     \toprule
      Model & ATOMIC & CWWV\\
      \midrule
      GPT2-L & 43.2 & 69.5 \\
      RoBERTa-L & 45.9 & 64.5 \\
      Human & 78.0 & 80.7\\
      \bottomrule
    \end{tabular}
  \end{center}
\end{table}

\subsection{Generating Fair and Informative Questions}

Alternatively, this result may be explained \cut{An alternative explanation for this result is indicated }by our human evaluation: not all automatically generated questions are fair and a subset has more than one correct answer, as a direct consequence of the inherent incompleteness of KGs. 
Besides being a disadvantage of automatic question generation, this finding points to a more substantial challenge: generating fair and informative multiple-choice questions is not yet well-understood. \cut{In an independent annotation study, the authors found that out of 200 questions from the \csqa~dataset that were answered incorrectly by  HyKAS~\cite{ma-etal-2019-towards}, \filip{number here + anonymize system} had more than one correct answer according to human annotators.} Adversarial strategies yield more plausible candidates than random sampling, making the task less fair; yet, fully relying on random sampling would generate distractors that are trivially discernible from the correct option. Balancing between fairness and informativeness, thus, is essential for multiple-choice question generation. Our empirical evidence suggests that it could be achieved by a mixed approach, where part of distractors is generated randomly, and part by adopting suitable adversarial strategies.\footnote{Question formulation is another challenge. Template-based questions may be trivially easy for LMs to solve, as discussed in \url{https://cs.nyu.edu/faculty/davise/papers/CYCQns.html}.}


\cut{\subsection{Scoring function}
While our scoring function already achieves strong performance across different tasks, we recognize that there are other scoring functions that may further improve the scores. For example, \cite{tamborrino-etal-2020-pre} showed that for masked language models, one can mask our N consecutive tokens at a time and to compute the score for sequences. The combination of scores computed when N=1,2,3... works better than N=1. However, such method also implies larger computational cost in both pretraining and evaluation stage. Since our main goal is not to achieve the SOTA performance, we leave the exploration of better scoring function for future work. \filip{this is useful but perhaps we can turn it into 'Limitations of the Framework' and discuss other things that we havent done yet}}

%% file: 6-conclusion.tex
\section{Conclusions}

While zero-shot QA is gaining focus, no study so far has investigated systematically the dependency between the task, the technique, and any additional knowledge used. To address this gap, this paper proposed a framework for zero-shot QA by pretraining LMs on artificial data created based on KGs. We studied the interplay between five knowledge sources, four data generation strategies, two LMs, two training regimes, and five tasks. We put forward seven hypotheses, which guided our experiments. We observed that language models benefit from pretraining on artificially created QA sets. Similar to prior work, we observed that the best performance is obtained by the RoBERTa model. Furthermore, combining knowledge sources leads to best performance, indicating that diverse relevant knowledge is desired for pretraining. However, this is conditioned on a good alignment between knowledge and the task. The training regime had a role too: marginal ranking is superior to vanilla language modeling, as it preserved better the structure of the task. Our analysis and human evaluation indicated that the generated questions were typically easier for humans than for language models, which is appropriate for commonsense questions.

Yet, the human evaluation revealed that a notable portion of the questions is nonsensical or difficult, hinting that automatically generating high-quality, informative commonsense questions is non-trivial and should be revised in subsequent work. Future work should also investigate the impact of this approach on knowledge injection systems~\cite{ma-etal-2019-towards} and graph relational networks~\cite{lin-etal-2019-kagnet}. It should also consider: (1) other, less structured knowledge sources, like WikiHow; (2) different distractor sampling strategies, e.g., based on iterative sampling \cite{niu2020self}; and (3) additional LM scoring functions, e.g., based on scoring sequences of tokens~\cite{tamborrino-etal-2020-pre}.